\documentclass[10pt,twocolumn,letterpaper]{article}

\usepackage[pagenumbers]{iccv} 

\definecolor{iccvblue}{rgb}{0.21,0.49,0.74}
\usepackage{scrextend}
\deffootnote{0em}{0em}{\thefootnotemark}
\usepackage[pagebackref,breaklinks,colorlinks,allcolors=iccvblue]{hyperref}
\usepackage{bbding,pifont}
\usepackage{breakurl}
\newcommand{\cmark}{\ding{51}}%
\newcommand{\xmark}{\ding{55}}%
\def\paperID{*****} 
\def\confName{ICCV}
\def\confYear{2025}

\title{VQualA 2025 Challenge on Engagement Prediction for Short Videos: Methods and Results}

\author{Dasong Li\textsuperscript{$*$} \quad Sizhuo Ma\textsuperscript{$*$} \quad Hang Hua\textsuperscript{$*$} \quad Wenjie Li\textsuperscript{$*$} \quad Jian Wang\textsuperscript{$*$} \quad Chris Wei Zhou\textsuperscript{$*$} \\
Fengbin Guan \quad Xin Li \quad Zihao Yu \quad Yiting Lu \quad Ru-Ling Liao \quad Yan Ye \quad Zhibo Chen \\
Wei Sun \quad Linhan Cao \quad Yuqin Cao \quad Weixia Zhang \quad Wen Wen \quad Kaiwei Zhang \\
Zijian Chen \quad Fangfang Lu \quad Xiongkuo Min \quad Guangtao Zhai \quad Erjia Xiao \\
Lingfeng Zhang \quad Zhenjie Su \quad Hao Cheng \quad Yu Liu\quad Renjing Xu \quad Long Chen \\
Xiaoshuai Hao \quad Zhenpeng Zeng \quad Jianqin Wu \quad Xuxu Wang \quad Qian Yu \quad Bo Hu \\
Weiwei Wang \quad Pinxin Liu \quad Yunlong Tang \quad Luchuan Song \quad Jinxi He \quad Jiaru Wu  \quad Hanjia Lyu
}

\begin{document}
\maketitle
\renewcommand{\thefootnote}{}
\footnotetext{$^{*}$Dasong Li (\textcolor{magenta}{dasongli@link.cuhk.edu.hk}), Sizhuo Ma (\textcolor{magenta}{sma@snap.com}), Hang Hua (\textcolor{magenta}{hhua2@cs.rochester.edu}),  Wenjie Li (\textcolor{magenta}{wenjie.li@snap.com}), \\Jian Wang (\textcolor{magenta}{jwang4@snap.com}), and Chris Wei Zhou (\textcolor{magenta}{zhouw26@cardiff.ac.uk}) are the challenge organizers of this challange.}
\footnotetext{The other authors are participants of the VQualA 2025 EVQA-SnapUGC: Engagement prediction for short videos Challenge.}
\footnotetext{The project page is \url{https://github.com/dasongli1/SnapUGC_Engagement/tree/main/ECR_inference}}
\begin{abstract}
This paper presents an overview of the VQualA 2025 Challenge on Engagement Prediction for Short Videos, held in conjunction with ICCV 2025. The challenge focuses on understanding and modeling the popularity of user-generated content (UGC) short videos on social media platforms. To support this goal, the challenge uses a new short-form UGC dataset featuring engagement metrics derived from real-world user interactions. This objective of the Challenge is to promote robust modeling strategies that capture the complex factors influencing user engagement. 
Participants explored a variety of multi-modal features, including visual content, audio, and metadata provided by creators. The challenge attracted 97 participants and received 15 valid test submissions, contributing significantly to progress in short-form UGC video engagement prediction.
\end{abstract} 
\section{Introduction}
With the rapid rise of social media, a growing number of content creators are sharing short videos that capture their daily lives on platforms like TikTok, Instagram Reels, YouTube Shorts, and Snapchat Spotlight. At the same time, a large share of users are spending significant amounts of time watching this type of content across these platforms.

Social media platforms receive a constant stream of newly published short videos. The effective dissemination of newly published videos remains a core objective for social media platforms. Recommending high-quality User Generated Content (UGC) videos enhances viewer engagement and consequently encourages content creators, especially novice creators. The effective dissemination of newly published videos remains a core goal of social media platforms. However, owing to their limited user reactions, accurate recommendation of such \emph{cold-start items} is usually a challenge. Typically, platforms would present each new video to a restricted number of users, such as one hundred. The latent popularity of each video is estimated based on the engagement metrics such as watch times from these initial users, serving as a basis for further recommendations. The cold start problem \cite{cold_start1,cold_start2,cold_start3,cold_start4} arises from the sampling bias in such limited initial interactions, resulting in noisy and inaccurate predictions of recommendation extents. Additionally, this conventional approach can result in time-sensitive short videos not being broadcast promptly, causing them to miss critical attention. Furthermore, emerging creators may struggle to gain sufficient visibility and recommendations, limiting their potential impact. Content creators may also face delays in gauging their videos' popularity, slowing their adjustments based on viewer feedback and thus discouraging them from posting more quality content. 
Consequently, an ineffective cold-start process may creates a negative feedback loop within the ecosystem, hindering the recommendation of high-quality videos to users, especially for some small-size or mid-size social media platforms.



\begin{figure*}[!t]
  \centering
    \small
    \begin{center}
    \setlength{\tabcolsep}{1pt}
    \begin{tabular}{@{} c c c c @{}}
    \includegraphics[width=.245\linewidth]{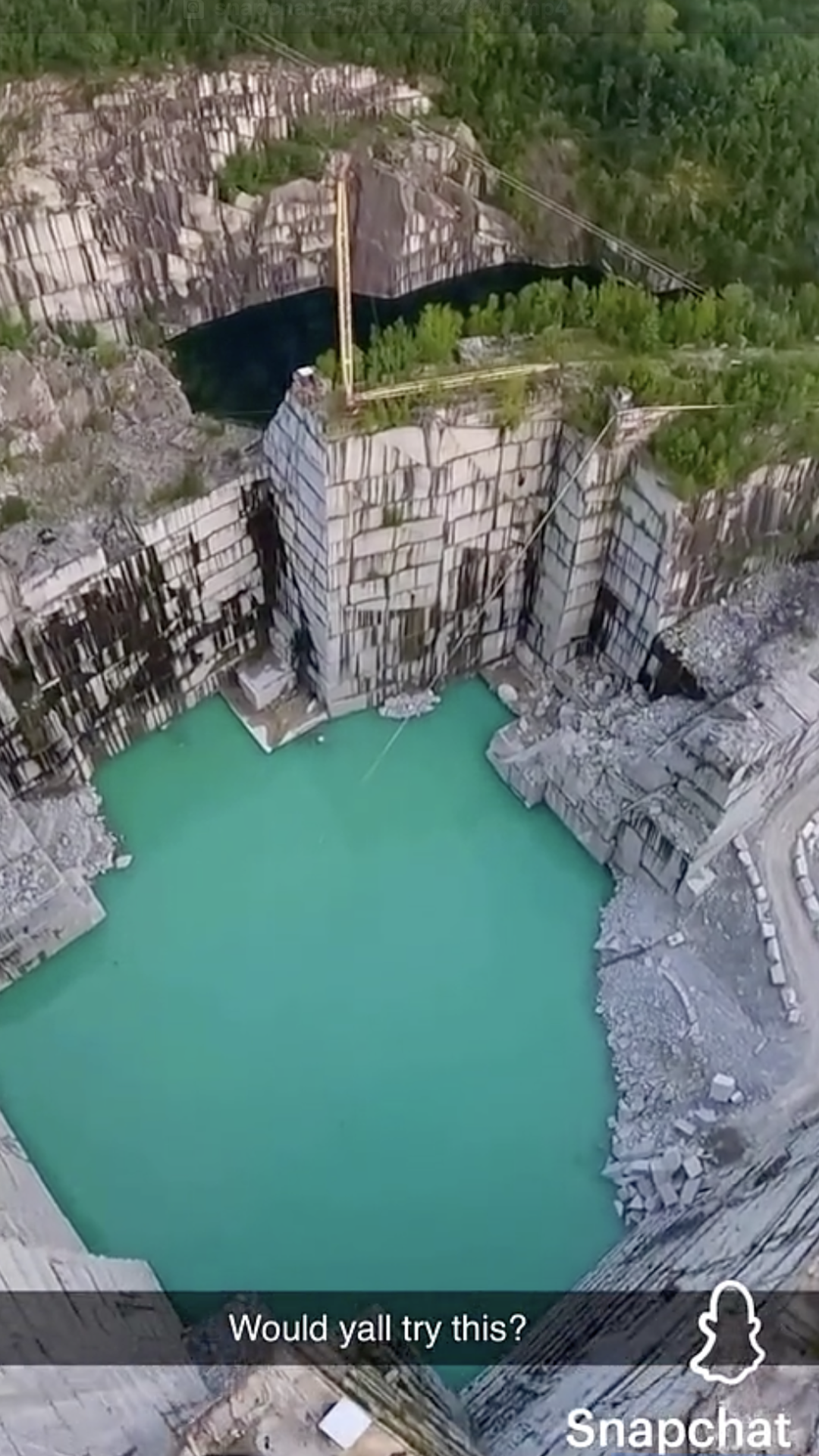} &
    \includegraphics[width=.245\linewidth]{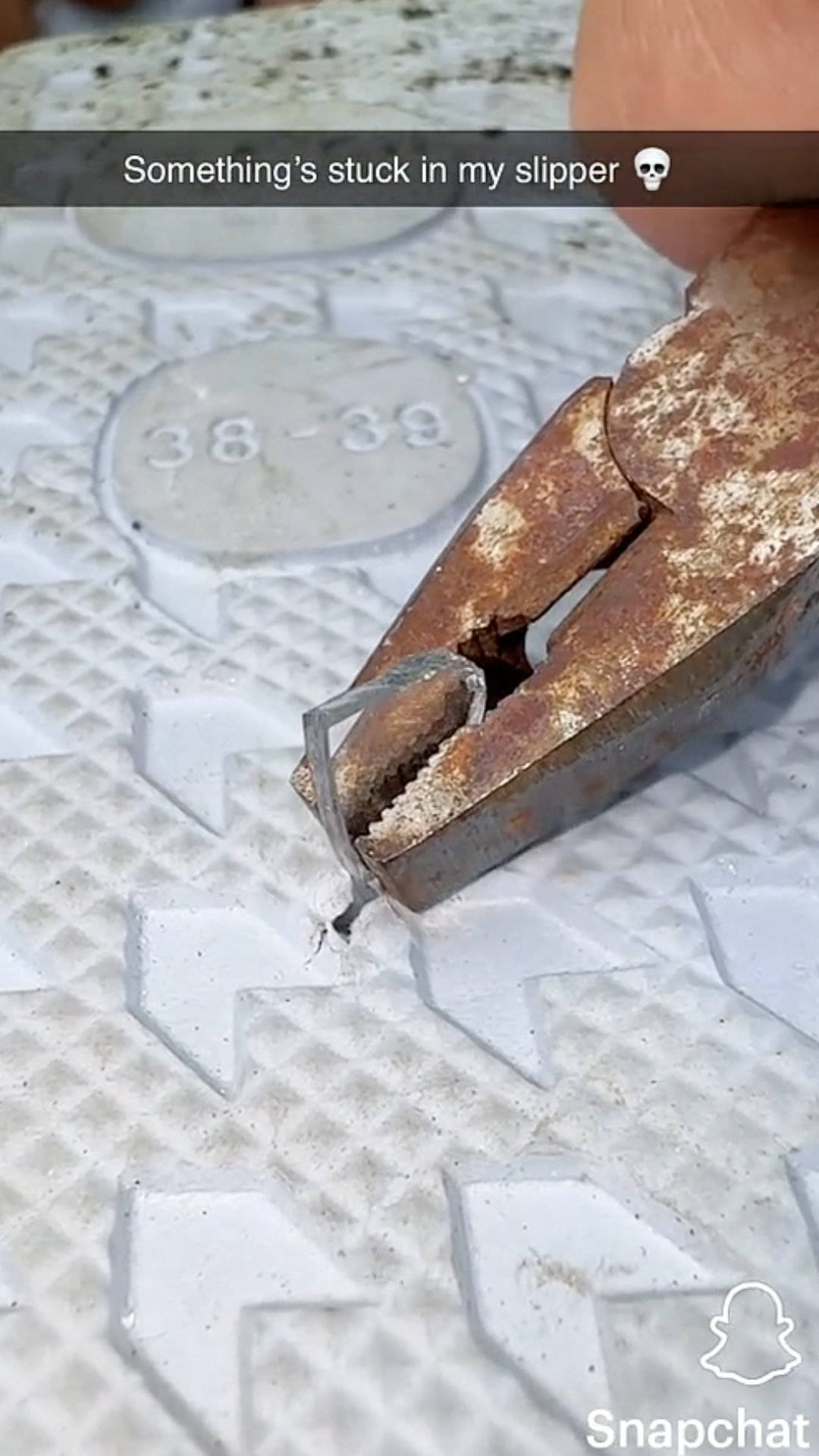} &
    \includegraphics[width=.245\linewidth]{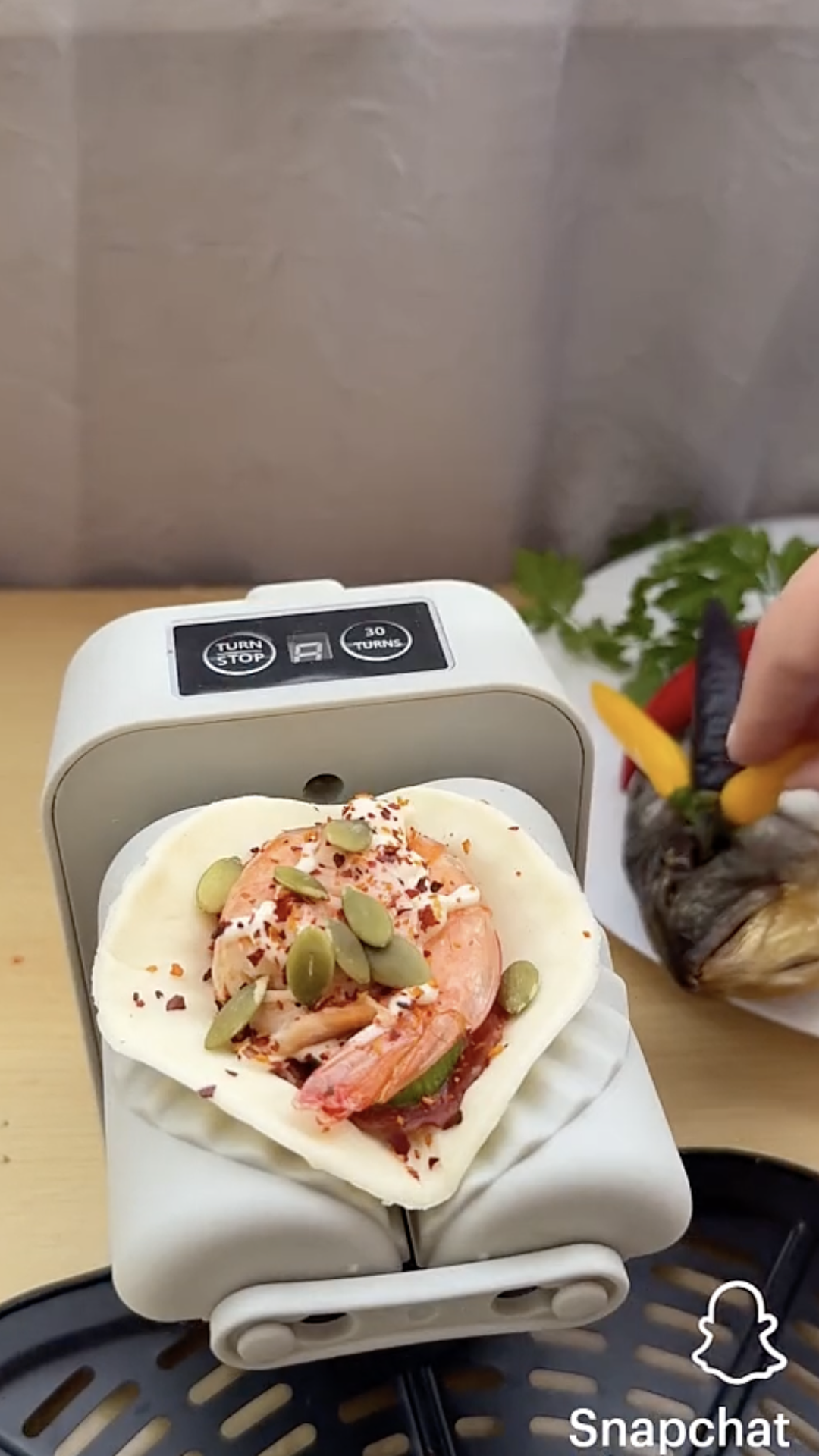} &
    \includegraphics[width=.245\linewidth]{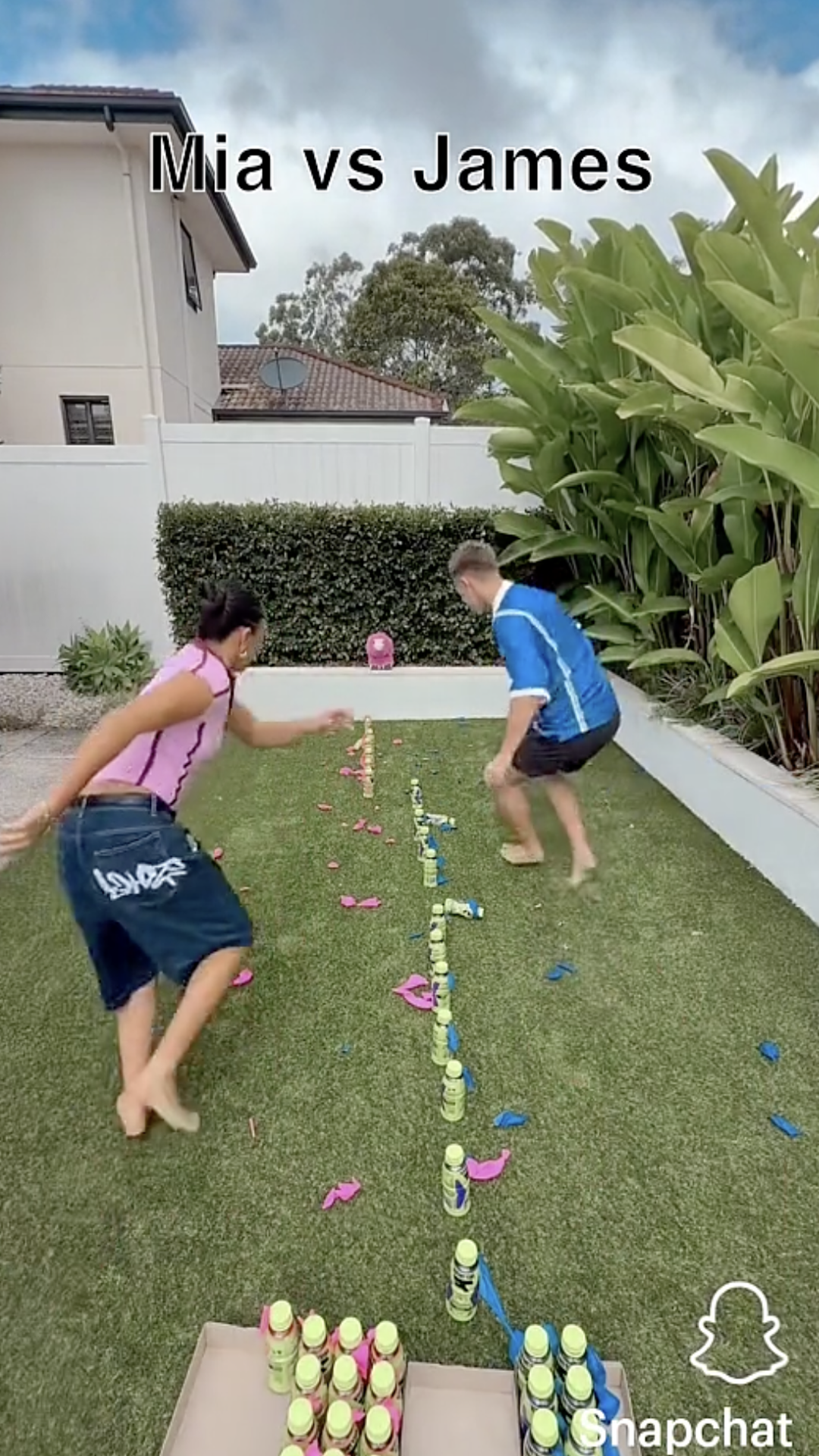} \\
    \end{tabular}
    \end{center}
    \caption{Sample frames of the short videos in SnapUGC dataset~\cite{li_engagement_eccv_2024}.}
    \label{fig:sample}
\end{figure*}

\begin{table*}[t]
  \centering
  \small
  \setlength{\tabcolsep}{10pt}
  \begin{tabular}{l|ccc|cc}
    \toprule
    & \multicolumn{3}{c|}{Multi-Modal Content} &  \multicolumn{2}{c}{Metrics} \\
     & Video & Audio & Text &  Annotators number & Metric Sources \\ \midrule
    VQA datasets~\cite{kv1k, ytugc, vqc, uvq, patch-vq}& \cmark & \xmark & \xmark & $\leq$ 40 & Labeling Scores \\
    Our dataset~\cite{li_engagement_eccv_2024} & \cmark & \cmark & \cmark & $\geq$ 1000 & Real User Interactions \\ 
    \bottomrule
  \end{tabular}
  \caption{We provide a detailed comparison with the VQA datasts. Our dataset contains multi-modal content to better measure the quality of videos. Moreover, our metrics are derived from thousands of real-world user interactions.}\label{tab:dataset1}
\end{table*}

A potential method for predicting engagement levels from video content is through user-generated content (UGC) video quality assessment (VQA). UGC VQA methods can be broadly classified into three categories based on the availability of reference information: full-reference~\cite{sun2021deep, LPIPS}, reduced-reference~\cite{soundararajan2012video,ma2012reduced}, and no-reference approaches~\cite{sun2022deep,tu2021ugc,internvqa,yu2024video}. The previous learning-based VQA methods~\cite{patch-vq,mlsp,vsfa,MD-VQA,uvq,additional_cite1,additional_cite2,additional_cite3,chen2025diffvqa} extract deep features via pre-trained models~\cite{he2016residual,efficientnet_v2,imagenet,resnet3d,k400data} and utilize these features to predict the MOS scores.
With the emergence of large language models (LLMs) and large multimodal models (LMMs), recent studies~\cite{wu2023qalign,lu2025qadapt,wu2024qinstruct} leverage their reasoning and interpretability capabilities to enhance the interactivity and explainability of VQA frameworks.


Despite the advancements in UGC VQA methods, Li et al. \cite{li_engagement_eccv_2024} demonstrated that VQA models \cite{uvq, patch-vq, Wu_2023_ICCV} trained on existing VQA datasets \cite{kv1k, ytugc, vqc, uvq, patch-vq} struggle to predict the popularity of short videos. This indicates that the mean opinion scores (MOS) annotated by small groups of human raters in video quality assessment datasets show a poor correlation with the popularity levels of these videos. This discrepancy may arise from the biases inherent in subjective MOS scores, which are influenced by the diverse preferences and limited participation of raters. As a result, these scores may not accurately reflect a video's appeal to its broader audience, as assessed by metrics like average watch time.
Furthermore, while VQA methods primarily focus on video visuals, short video engagement can be affected by additional factors such as background music, content category, and titles. Therefore, engagement prediction and video quality assessment are distinct tasks due to the differing nature of their datasets.

\begin{figure*}[!t]
    \small
    \begin{center}
    \setlength{\tabcolsep}{0.5pt}
    \begin{tabular}{@{} c c @{}}
    \begin{tabular}{@{} c @{}}
               \includegraphics[width=0.03\linewidth]{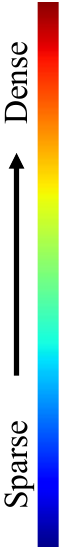} 
    \end{tabular} &
    \setlength{\tabcolsep}{0pt}
    \begin{tabular}{@{} c c c @{}}
    \includegraphics[width=.32\linewidth]{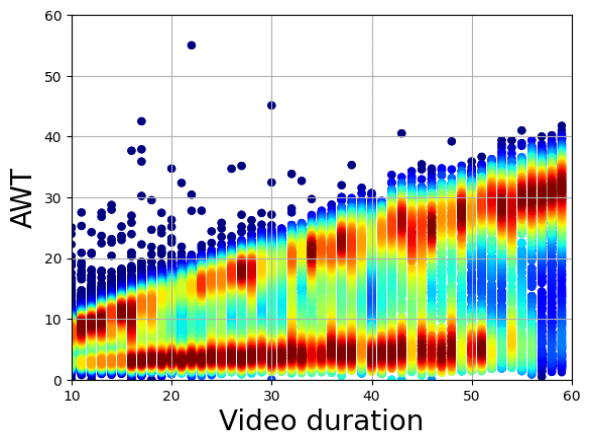} &
         \includegraphics[width=.32\linewidth]{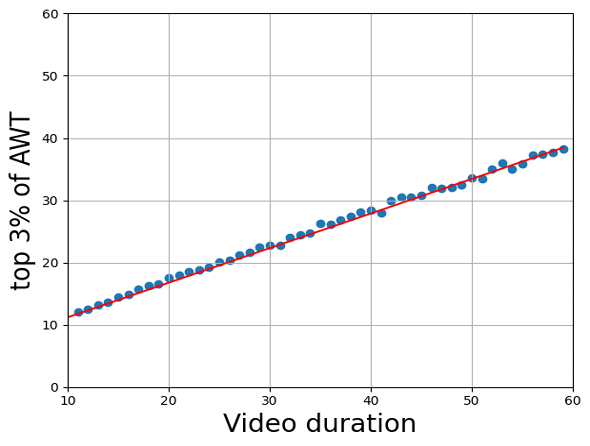} &
         \includegraphics[width=.32\linewidth]
         {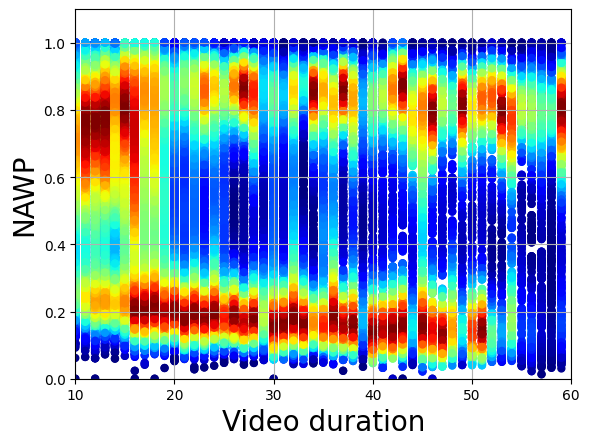} \\
         (a) Average watch time (AWT). & (b) Fitting top 3 \% of AWT. & (c) NAWP \\
         \includegraphics[width=.32\linewidth]
         {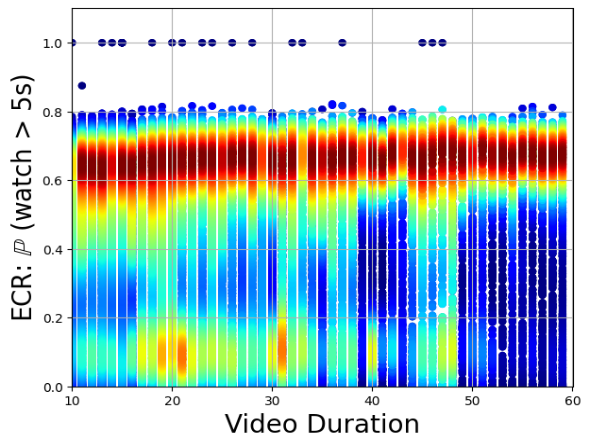} & 
         \includegraphics[width=.32\linewidth]
         {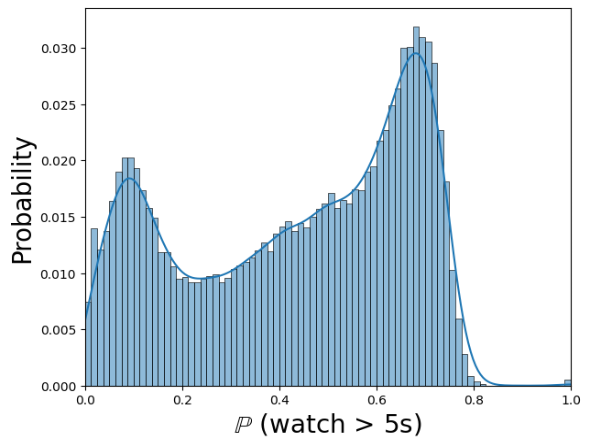} & 
     \includegraphics[width=.32\linewidth]{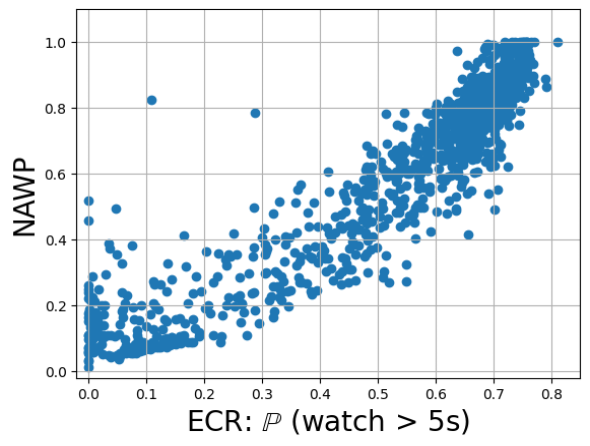} \\
        (d) Engagement continuation rate (ECR) & (e) Distribution of ECR & (f) Correlation between ECR and NAWP \\
        
    \end{tabular}
    \end{tabular}
    \end{center}
    \caption{(a), (d): The distributions of average watch time (AWT) and engagement continuation rate (ECR), respectively. ECR, calculated as the probability of watch time exceeding 5 seconds: $\mathbb{P}$ (watch \textgreater~5s), is more duration-independent. 
    (b): We fit top 3\% of average watch times to derive a universal metric for videos of different durations. (d): Further normalization of the average time is achieved by fitting a line, resulting in the normalized average watch percentage (NAWP). 
    A color mapping is used to encode the distribution densities in (a), (c), (d). (e): Distributions of ECR. ECR follows a bimodal distribution, reflecting the unique property of user's swiftly skipping uninteresting videos or spend relative longer time on their interesting videos in short videos platforms. (f): The strong correlation between ECR and NAWP.}
  \label{fig:analysis}
\end{figure*}
To address these limitations, we introduce a large-scale SnapUGC dataset of publicly accessible short videos on Snapchat Spotlight directly model the engagement levels~\cite{lecture_dataset,Wu_2018_beyond,kuaishou}. Unlike prior datasets, SnapUGC leverages real engagement data from over 2,000 users to mitigate the bias introduced by small-scale subjective annotation. We introduce two robust metrics to quantify engagement:
\begin{enumerate}
    \item \textbf{Normalized Average Watch Percentage (NAWP):} Measures overall user engagement normalized across videos of varying lengths. 
    \item \textbf{Engagement Continuation Rate (ECR):} Represents the probability that a viewer watches beyond the initial 5 seconds, indicating the video's ability to capture attention early on.
\end{enumerate}
NAWP provides an indication of the overall engagement level for videos with different durations and ECR assesses whether the video's outset is captivating enough to retain viewers' interest in continuing to watch. 
These metrics are computed in aggregate to ensure individual user privacy—no personal information or user histories are included in the dataset.

To further advance research on user engagement modeling for short-form videos, we are organizing the Engagement Prediction for Short Videos Challenge (EVQA) as part of the VQualA 2025 Workshop @ ICCV. This challenge aims to establish a practical and comprehensive benchmark for predicting viewer engagement, with \textbf{a specific focus on Engagement Continuation Rate (ECR) prediction} as the core task, selected for simplicity and clarity.
We are grateful to participants from both academia and industry for contributing to this shared goal of advancing short-form video quality assessment and engagement prediction.

This Challenge is one of VQualA 2025 Workshop associated challenges on: ISRGC-Q-image super-resolution generated content quality assessment~\cite{isrgcq2025iccvw}, FIQA-face image quality assessment~\cite{ma2025fiqa}, Visual quality comparison for large multimodel models~\cite{zhu2025vqa}, GenAI-Bench AIGC video quality assessment~\cite{genai-bench2025iccvw}, and Document Image Quality Assessment~\cite{diqa2025iccvw}.




\begin{table*}[]
\resizebox{\textwidth}{!}{
\begin{tabular}{c|c|c|ccc|c|c}
\hline
Rank & Team name          & Team leader            & Final Score & SROCC & PLCC  & Features & Large Multi-modal Models \\ \hline 
-   & Baseline & - &  0.660 & 0.657 & 0.665 & Multi-Modal & - \\ \hline
1   & ECNU-SJTU VQA      & Wei Sun            & 0.710       & 0.707 & 0.714  & Multi-Modal & Video-LLaMA (1.7B), Qwen2.5-VL (7B) \\
1*  & IMCL-DAMO          & Fengbin Guan         & 0.698      & 0.696 & 0.702 & Multi-Modal & Qwen2.5-VL (7B) \\
3   & HKUST-Cardiff-MI-BAAI    & Xiaoshuai Hao      & 0.680       & 0.677 & 0.684  & Visual Only & - \\
4    & MCCE & Zhenpeng Zeng                & 0.667       & 0.666 & 0.668  & Multi-Modal & - \\
4*    & EasyVQA            & Bo Hu            & 0.667       & 0.664 & 0.671  & Multi-Modal & - \\
6    & Rochester        &      Pinxin Liu       & 0.449       & 0.405 & 0.515   & Multi-Modal & Skywork-VL-Reward (7B) \\ \hline
7    & brucelyu          &      Hanjia Lyu     & 0.441       & 0.439 & 0.444 & Textual Only & - \\ \hline
\end{tabular}
}
\caption{Result of engagement prediction challenge.}
\label{tab:results_1}
\end{table*}

\section{Challenge Dataset}
\subsection{SnapUGC Datasset Collection}
To precisely model the engagement levels of real UGC short videos, we first collect a large-scale short video dataset, named SnapUGC. Our dataset comprises 120,651 short videos, all of which were published on Snapchat Spotlight. For each video, we have curated corresponding aggregated engagement data derived from viewing statistics.
All short videos in our dataset have a duration ranging from 5 to 60 seconds. To mitigate sampling bias from small number of views, only short videos with view numbers exceeding 2000 are selected.
The dataset is notably diverse, encompassing a wide range of video types, including Family, Food \& Dining, Pets, Hobbies, Travel, Music Appreciation, Sports, etc. Several frames are shown in Figure~\ref{fig:sample}. We provide a comprehensive comparison with traditional VQA datasets in Table~\ref{tab:dataset1}. 
The dataset is shown in the following:
\begin{enumerate}
    \item \textbf{Train set:} 106,192 short-form videos. Each video is accompanied with title and descriptions provided by creators.
    \item \textbf{Validation set:} 6000 short-form videos. Each video is accompanied with title and descriptions provided by creators.
    \item \textbf{Test set:} 8,459 short-form videos. Each video is accompanied with title and descriptions provided by creators.
\end{enumerate}

\subsection{Engagement Metrics}

Average watch time (AWT) is a naive and common metric to measure viewer engagement. However, AWT faces limitations when comparing videos of different durations. We first analyze the distribution and drawback of AWT, and then propose normalized average watch percentage (NAWP) as a novel engagement metric. Recognizing that users swiftly navigate through uninteresting content but persist in watching engaging videos, we introduce an additional metric: engagement continuation rate (ECR). Calculated for each video, this metric represents \emph{the proportion of viewers who watched the video for at least 5 seconds}. It serves as an indicator of a video's ability to captivate viewers at the beginning. Unlike Kim \etal \cite{MOOC} measuring entire videos' dropout probability, ECR focuses on the contents of first several seconds, which determines whether the users would continue to watch and substantially affects watch times. 

\noindent\textbf{Average watch time (AWT).}
We analyze average watch times (AWT) of various video durations $d$ in Figure~\ref{fig:analysis}(a). Importantly, the distributions of AWT vary for different video durations, showing diverse user engagement patterns. Therefore comparing the popularity of short videos with different durations using AWT is challenging.

\noindent\textbf{Normalized average watch percentage (NAWP).}
We introduce a straightforward metric called normalized average watch percentage (NAWP) to provide a generalized measure for videos with different durations. 
It is observed in Figure~\ref{fig:analysis}(a) that the largest values under different durations align with a linear trend. Based on the observation, we make the assumption that videos with top 3\% of highest AWT, regardless of their durations, are equally most popular, while videos with an average watch time of 0 seconds are deemed the least popular.
The maximum average watch time $f_{\text{max}}(d)$ for most popular videos and minimum average watch time $f_{\text{min}}(d)$ for the least popular videos can be modeled by two linear functions:
\begin{equation}
   f_{\text{max}}(d) = \alpha \times d + \beta;~f_{\text{min}}(d) = 0.
\end{equation}
$f_{\text{max}}(d)$ is shown in Figure~\ref{fig:analysis}(b). The NAWP for any video of $d$ seconds, with average watch time $t$ is derived through normalization between $f_{\text{min}}(d)$ and $f_{\text{max}}(d)$:
\begin{equation}
    \text{NAWP}(\text{AWT},d) = \min\left(\frac{\text{AWT} - f_{\text{min}}(d)}{f_{\text{max}}(d)-f_{\text{min}}(d)},1\right). 
    \label{eq:clip}
\end{equation}
The relationship between the video duration and NAWP is depicted in Figure~\ref{fig:analysis}(c). The NAWP falls within the range of [0, 1] and NAWP of videos with top 3\% average watch time is set to be 1.

\noindent\textbf{Engagement continuation rate (ECR).}
As shown in Figure~\ref{fig:analysis}(e), engagement continuation rate (ECR), calculated as $\mathbb{P}$ (watch \textgreater 5s), demonstrates stable behavior across different video durations. The majority of values fall within the range of [0, 0.8].
\begin{figure*}[!t]
  \centering
    \includegraphics[width=0.98\linewidth]{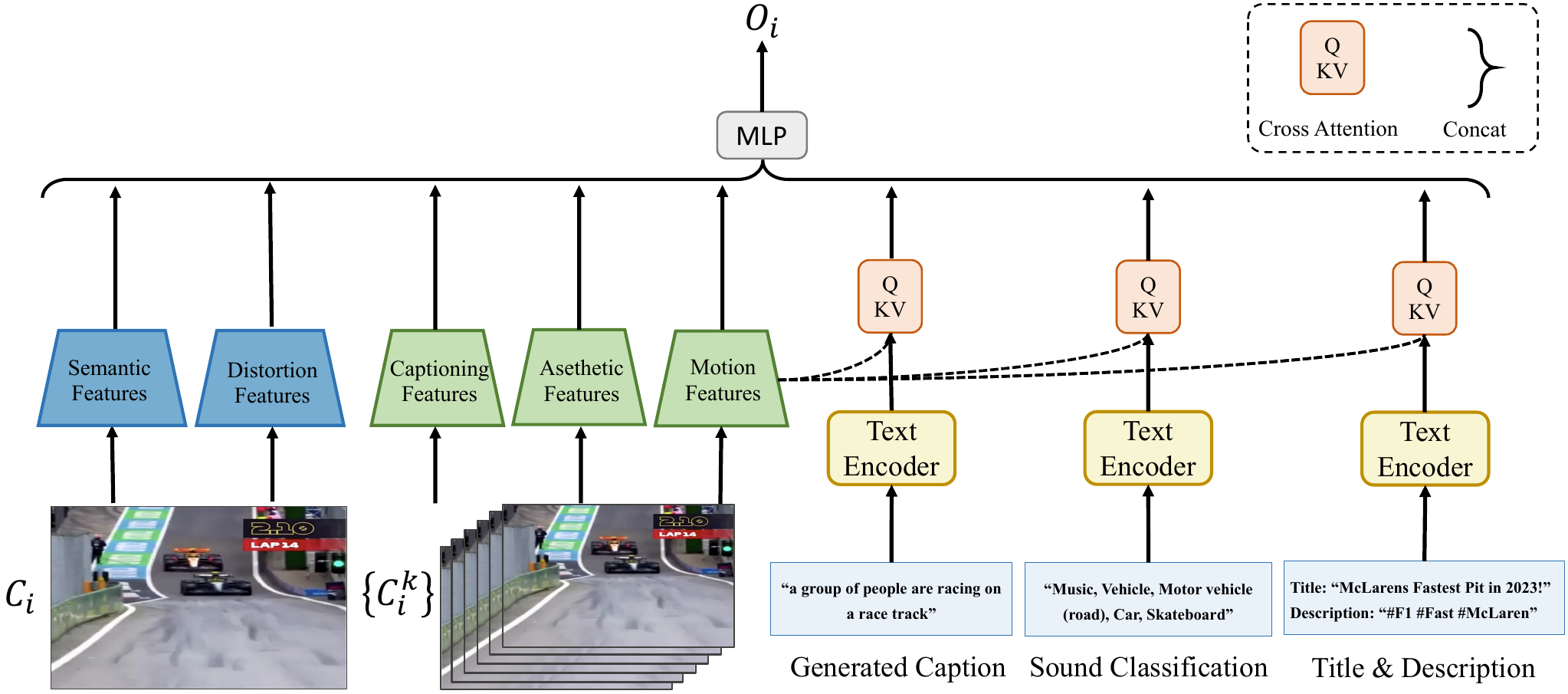} 
    \caption{The overview framework of baseline.}
    \label{fig:investigate1}
    \includegraphics[width=0.98\linewidth]{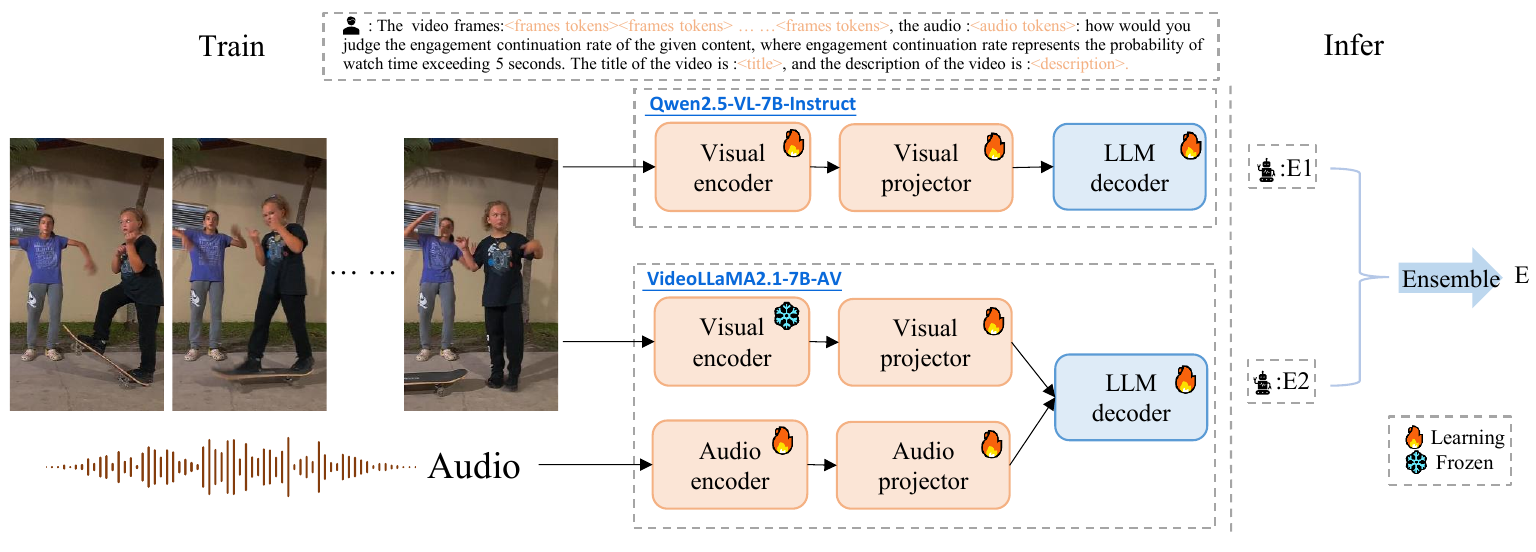} 
    \caption{The overview framework provided by Team ECNU-SJTU VQA.}
    \label{fig:team1}
\end{figure*}

\noindent\textbf{ECR for the Challenge.}
the $\alpha$ and $\beta$ in NAWP may vary across different datasets or different platforms. The ECR and NAWP are observed to have a strong correlation of 0.928 in Figure~\ref{fig:analysis}(f). Therefore, we select ECR as the metrics in EVQA Challenge. To protect the private information of creators, the ECR used in this challenge is derived from normalizing the ranking of real ECR.

\section{Challenge Results}
The challenge results are summarized in Table~\ref{tab:results_1}, including the performance of all teams that submitted their fact sheets. As this is a novel task involving multi-modal features, we did not impose restrictions on model size in order to explore the upper bound of model capacity. We provide a baseline model achieving an SROCC of 0.660 and a PLCC of 0.657, based on the approach proposed by Li \etal~\cite{li_engagement_eccv_2024}.

The teams with top performances including ECNU-SJTU VQA, ICML-DAMO, HKUST-Cardiff-MI-BAAI, MCCE(MCCE (Media Convergence and Communication Experimental)), EasyVQA achieved excellent results in both PLCC and SROCC, exceeding our baseline. 
Among them, ECNU-SJTU VQA and ICML-DAMO demonstrated the most significant improvements over the baseline. Given their competitive performance, these two teams are recognized as co-first place. As shown in Table~\ref{tab:results_1}, large multimodal models were widely adopted to boost performance.
Interestingly, the team brucelyu achieved reasonable performance using only textual features (\eg, title, description, and music classification), highlighting that non-visual information can also meaningfully contribute to predicting user engagement with short videos.

\begin{figure*}[!t]
  \centering
    \includegraphics[width=0.9\linewidth]{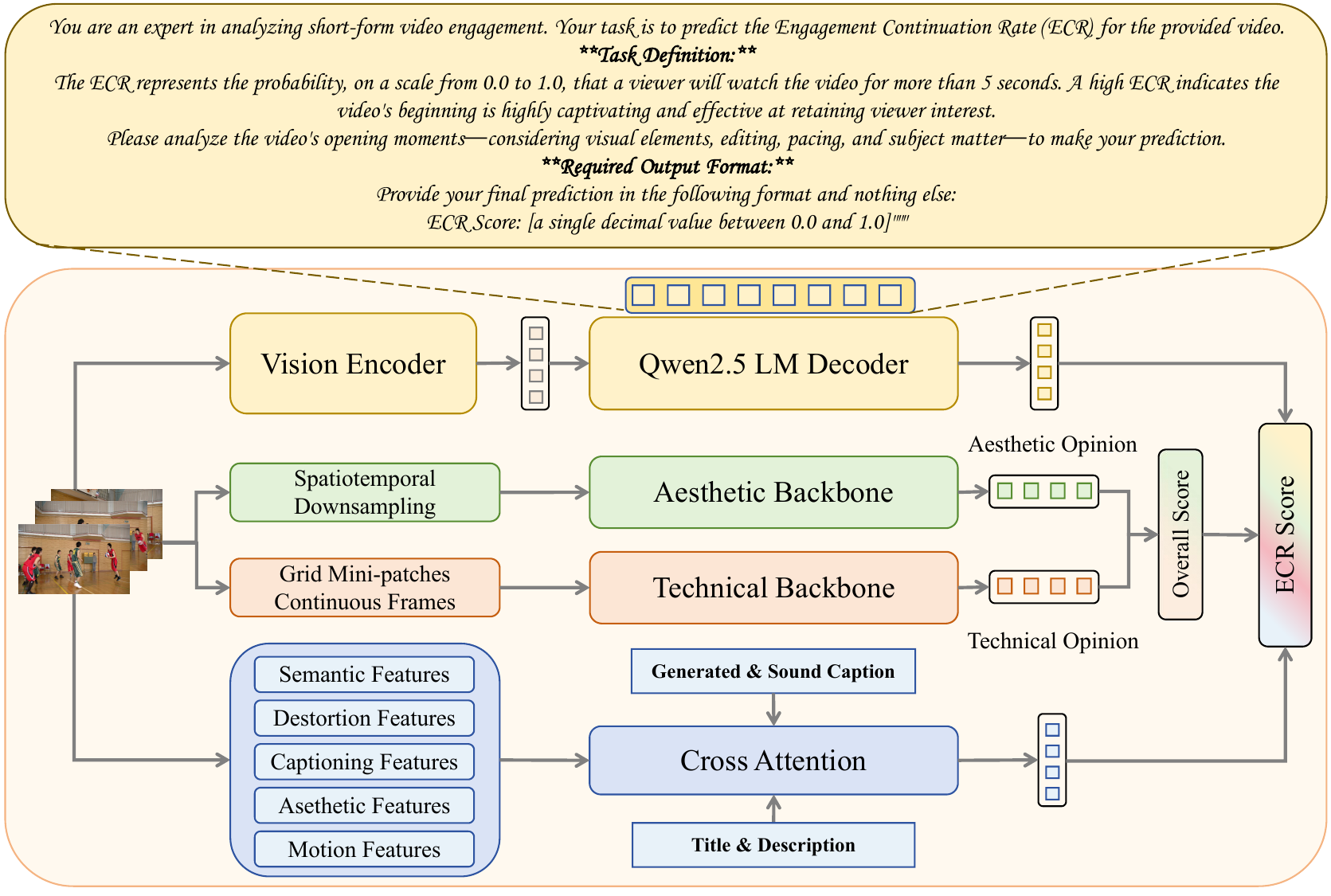} 
    \vspace{-0.3cm}
    \caption{The overview framework provided by Team IMCL-DAMO.}
    \label{fig:team2}
\end{figure*}

\section{Teams and Methods}
\subsection{Baseline}
The provided baseline, built on the Li \etal~\cite{li_engagement_eccv_2024}, are trained on the ECR prediction. A comprehensive set of multi-modal features, including per-frame semantic features~\cite{efficientnet_v2}, per-frame pixel-level distortion features~\cite{uvq} from different degradations~\cite{Li_2023_CVPR,Li2022Learning,Li2022efficient}, sound classification~\cite{tensorflow2015-whitepaper}, text descriptions from authors, video captioning~\cite{Xu2023mPLUG2AM}, are used. The framework of baseline is shown in Figure~\ref{fig:investigate1}.

\subsection{ECNU-SJTU VQA Team}
This approach utilizes an ensemble of Large Multimodal Models (LMMs)~\cite{sun2025engagement} for video quality score prediction. Specifically, they leverage two LMMs: Video-LLaMA2 \cite{damonlpsg2024videollama2}, a high-performance model tailored for audio-visual language understanding, and Qwen2.5-VL~\cite{bai2025qwen2}, a powerful model focused on general vision-language tasks.

For Video-LLaMA2~\cite{cheng2024videollama}, they provide the model with the first 8 (or 5) video frames, the audio track, and the associated video description text (including the video title and description; if unavailable, they use ‘None’ as the default input). To predict the Engagement Continuation Rate (ECR), they extract the hidden features from the last layer of Video-LLaMA2 and append a regression head. The regression head consists of a multilayer perceptron (MLP), which includes a dropout layer, a fully connected (FC) layer with 2048 neurons, a ReLU activation layer, and a final FC layer with a single neuron to predict the video engagement score.

For Qwen2.5-VL~\cite{bai2025qwen2}, they similarly provide the first 8 video frames and the corresponding video description text as input. For ECR prediction, they follow the original Qwen2.5-VL architecture and utilize the next-token output for regression. Finally, they ensemble the predictions from both models to obtain the final engagement score.

\begin{figure*}[!t]
  \centering
    \includegraphics[width=0.94\linewidth]{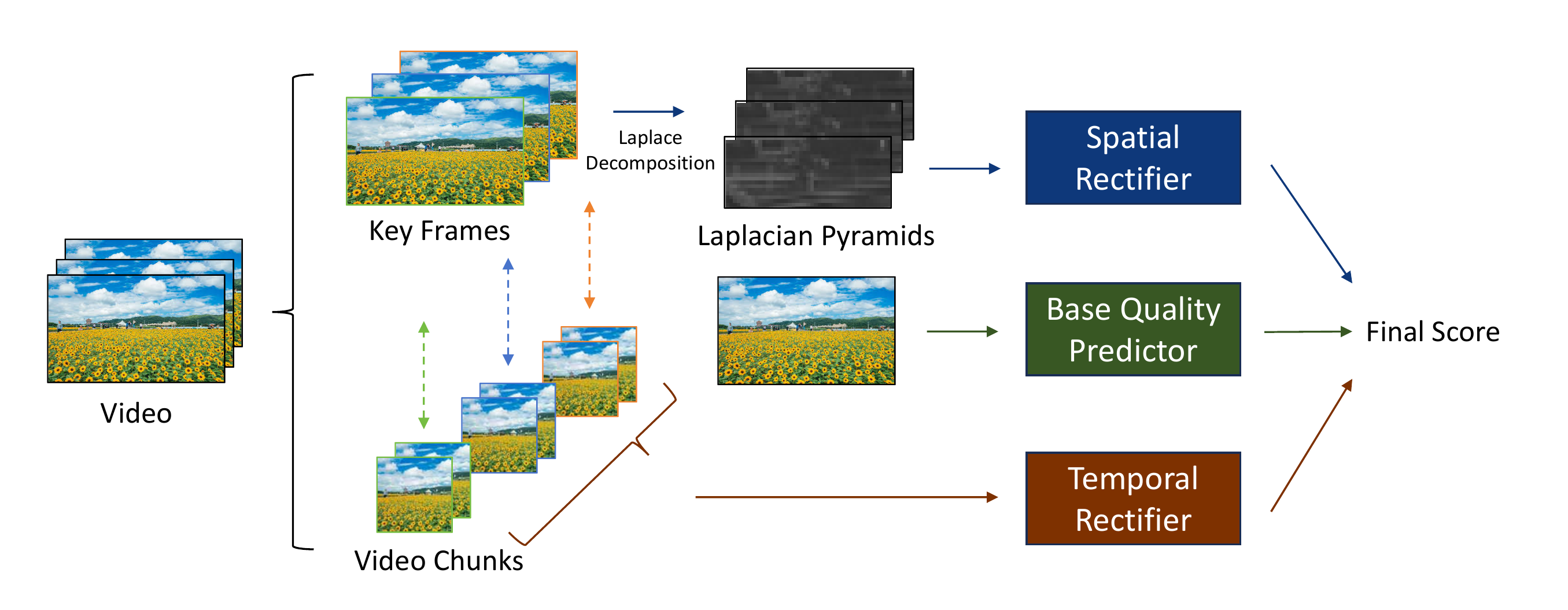} 
    \vspace{-0.5cm}
    \caption{The overview framework provided by Team HKUST-Cardiff-MI-BAAI .}
    \label{fig:team3}
\end{figure*}
\noindent\textbf{Training details.} For VideoLLaMA2.1-7B-AV~\cite{cheng2024videollama}, each input image is first resized to a global resolution of 384$\times$384, then divided into multiple 384$\times$384 patches using grid-based cropping and padding. These patches are jointly fed into the vision encoder. They finetune the model on 2 A800 GPUs with a batch size of 12 for one epoch. During training, the parameters of the vision encoder are frozen, while the remaining parameters are updated. The model is optimized using a learning rate of 5 $\times$ $10^{-5}$. Then they train VideoLLaMA2.1-7B-AV with different random seeds and number of input frames. For Qwen2.5-VL-7B-Instruct~\cite{bai2025qwen2}, they control the maximum number of image pixels to be image max pixels = 768$\times$28$\times$28 to ensure efficient memory usage. The model is trained on 8 A800 GPUs with a batch size of 16 for one epoch. Similar to LLaVA, we update all parameters, applying a learning rate of 2 $\times$ $10^{-6}$ to the vision encoder and 1 $\times$ $10^{-5}$ to the rest of the model.

\noindent\textbf{Testing details.} They evaluate our framework on the EVQA dataset using three models. For VideoLLaMA2.1-7B-AV, each image is directly resized to 384$\times$384 prior to inference. For the Qwen2.5-VL models, we apply the same preprocessing as used during training, constraining the maximum number of image pixels to 768$\times$28$\times$28. The final ensemble prediction is obtained by combining the outputs of three VideoLLaMA2.1-7B-AV models and one Qwen2.5-VL-7B-Instruct model.

\subsection{IMCL-DAMO Team}
This team combines three main branches: (1) a baseline model that extracts diverse multi-modal features to enhance video-content relevance; (2) fragment-based sampling leveraging DOVER~\cite{Wu_2023_ICCV}’s technical and aesthetic branches to capture multiple quality perspectives; and (3) a Qwen2.5-VL-7B~\cite{bai2025qwen2} branch utilizing full-parameter supervised fine-tuning to incorporate strong multi-modal priors for ECR prediction.

\noindent\textbf{Training details.} They used PyTorch and the Transformers library to implement all models. The training involved only the official competition training dataset. Full parameter fine-tuning was conducted for both the DOVER branch and the Qwen2.5-VL-7B model. Experiments were run on 4 NVIDIA A100 GPUs, requiring approximately $48\sim72$ hours of training time. They employed mixed precision training and standard optimizer configurations to improve efficiency.

\noindent\textbf{Testing details.}
Evaluation was performed on the official competition test set with a batch size of 1. No multi-scale or test-time augmentation strategies were applied. The baseline branch required over 12 hours of inference, the Qwen2.5-VL-7B branch about 6 hours, and the aesthetic/technical DOVER branches between 10–20 minutes.

\subsection{HKUST-Cardiff-MI-BAAI Team}
This team’s approach integrates three main components:
\begin{enumerate}
    \item \textbf{Base Quality Predictor:} This module takes a sparse set of spatially downsampled key frames as input and uses a pretrained Vision Transformer (ViT) \cite{dosovitskiy2020image} from CLIP~\cite{radford2021learning} to generate a scalar quality estimate.
    \item \textbf{Spatial Rectifier:} This component processes Laplacian pyramids of key frames at the original spatial resolution to compute scaling and shift parameters, which are used to refine the base quality score.
    \item \textbf{Temporal Rectifier:} This module processes spatially downsampled video chunks centered around key frames at the original frame rate, producing another set of scaling and shift parameters to further adjust the quality estimate.
\end{enumerate}
\begin{figure*}
 \centering
    \includegraphics[width=0.85\linewidth]{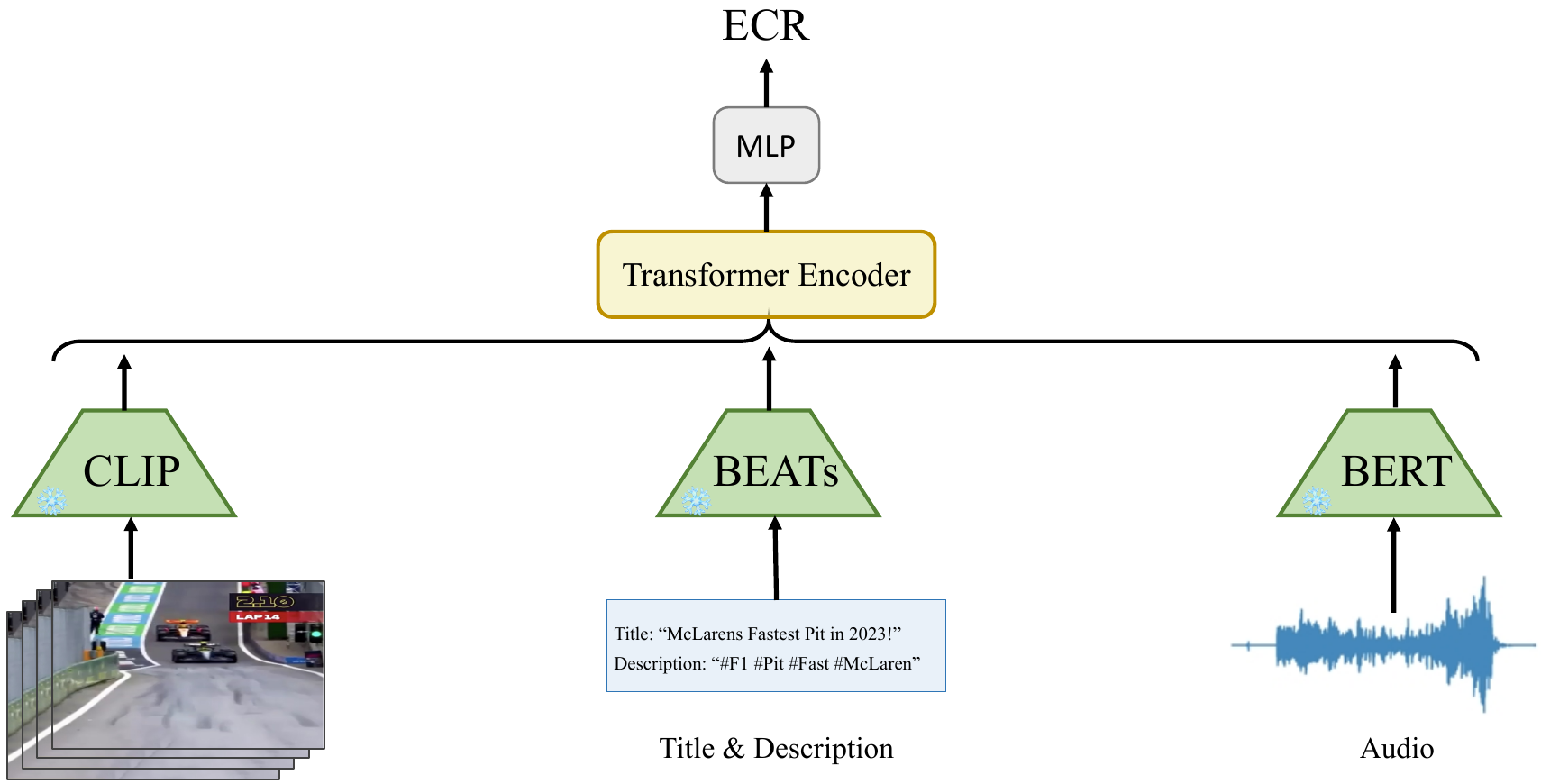} 
    \vspace{-0.3cm}
    \caption{The overview framework provided by Team EasyVQA.}
    \label{fig:team5}
\end{figure*}

\noindent\textbf{Training details.} They used PyTorch to implement all models. The training involved only the official competition training dataset. They fine-tune the model with a batch size of 32 for two epochs. Spatial and temporal rectifiers are randomly dropped out during training with probabilities of 0.1. The model is optimized by Adam optimizer using a learning rate of 5 $\times$ $10^{-5}$.

\subsection{MCCE (Media Convergence and Communication Experimental) Team}
This team enhanced the provided baseline by incorporating multi-modal fusion techniques, novel perceptual features, and temporal aggregation of video features:
\begin{enumerate}
    \item \textbf{Novel Image Features Fusion:} They generate attention weights from semantic features to dynamically adjust the fusion ratio of distortion features, placing greater emphasis on distortion features in key frames.
    \item \textbf{Novel Perception Features:} A more advanced visual encoder (perceptual encoder) is introduced to extract richer video perception features, enabling more comprehensive video representation learning.
    \item \textbf{Motion Features Temporal Aggregation:} They employ a global-local temporal aggregation mechanism to capture dynamic changes in short videos, such as action intensity and transition rhythm.
\end{enumerate}
The team also achieves a $3\sim5\times$ speedup in training by leveraging advanced memory optimization, distributed computing, and intelligent resource management. Key techniques include LRU-based memory management and GPU memory-aware batch size adjustment.

\subsection{EasyVQA Team}
This team focuses on video content and leverages multi-modal information—visual, textual, and auditory—to enhance representation learning. They utilize CLIP~\cite{radford2021learning} to extract features from video frames, BERT \cite{devlin2019bert} for textual features from video titles, and BEATs \cite{chen2022beats} for audio features. As a preprocessing step, video data is processed based on the ECR metric: the first frame of each second within the first 5 seconds of a video is extracted as a key frame. These frames are resized to 224$\times$224 pixels and passed through CLIP’s visual encoder. Simultaneously, the audio and title text are processed to obtain audio tokens and text tokens, respectively. The model is trained to distinguish among the different modalities. To this end, modality-specific embeddings are added to the features of each modality, and positional encodings are applied to the video tokens. All tokens, including visual, textual, and auditory, are then concatenated and passed through a Transformer Encoder to generate a unified video feature representation. Finally, an MLP head predicts the ECR value.

\subsection{Rochester Team}
This team employs the pretrained Skywork-VL-Reward-7B vision-language model~\cite{wang2025skyworkvlrewardeffectivereward} to predict a continuous quality score in the range [0, 1] for input videos. Sixteen frames are evenly sampled from each video and processed using the frozen Skywork-VL backbone, followed by a lightweight regression head that outputs the final score. Thanks to the strong multimodal foundation of the pretrained model and the simplicity of the regression head, the approach achieves high efficiency and strong alignment with human evaluations.

\noindent\textbf{Training details.}
The model is implemented using PyTorch and trained using the AdamW optimizer with weight decay. The initial learning rate is set to 1$\times$ $10^{-6}$ and decayed via cosine annealing across 5 epochs. Training was conducted using mixed-precision (FP 16) over 4 NVIDIA H100 GPUs for a total duration of 32 hours. They use the competition-provided dataset exclusively, without any external data. Frame sampling enables full-resolution input while maintaining computational efficiency. Gradient accumulation was used to simulate larger batch sizes.

\noindent\textbf{Testing details.}
During inference, 16 evenly sampled frames from the video are passed through the frozen Skywork-VL-Reward-7B \cite{wang2025skyworkvlrewardeffectivereward} model. The features are then fed into a regression head to produce a single score between 0 and 1.

\subsection{brucelyu17 Team}
This team uses the XGBoost regression model \cite{chen2016xgboost}, implemented via the xgboost Python package, to predict short video engagement levels using only textual features. The input comprises semantic representations derived from both textual and categorical data.

For textual features, the team applies the pre-trained Sentence-BERT model (all-MiniLM-L6-v2) \cite{reimers2019sentence} to convert video titles and captions into dense embeddings. These embeddings capture the contextual semantics of the text and provide compact, informative representations for modeling.

For categorical features, they incorporate background music category information. Initially encoded using multi-hot vectors, these features are further transformed into semantic embeddings to better reflect relationships among music categories. 

To optimize model performance, they perform a grid search over key XGBoost hyperparameters and select the configuration that achieves the lowest Root Mean Squared Error (RMSE) on the validation set. The hyperparameter search space includes maximum tree depth (3, 6), learning rate (0.1, 0.05, 0.01), and subsample ratio (0.8, 1.0). Each parameter combination is evaluated using early stopping to avoid overfitting. The final model is trained using the best configuration and evaluated on a held-out test set.

\clearpage

{
    \small
    \bibliographystyle{ieeenat_fullname}
    \bibliography{main}
}
\appendix
\section{Teams and Affiliations}

\label{sec:team}
\subsection*{VQualA 2025 EVQA Track Organizers}
\noindent\textbf{Members:} Dasong Li$^1$ (\url{dasongli@link.cuhk.edu.hk}), Sizhuo Ma$^2$ (\url{sma@snap.com}), Hang Hua$^3$ (\url{hhua2@cs.rochester.edu}), Wenjie Li$^2$ (\url{wenjie.li@snap.com}), Jian Wang$^2$ (\url{jwang4@snap.com}) and Chris Wei Zhou$^4$ (\url{zhouw26@cardiff.ac.uk})\\
\textbf{Affiliations:}\\
$^1$The Chinese University of Hong Kong. \\
$^2$Snap Inc.\\
$^3$University of Rochester.\\
$^4$Cardiff University.

\subsection*{IMCL-DAMO}
\textbf{Members:} Fengbi Guan$^{1,2}$ (\url{guanfengbin.gfb@alibaba-inc.com}), Xin Li$^1$ (\url{xin.li@ustc.edu.cn}), Zihao Yu$^{1,2}$ (\url{zuhe.yzh@alibaba-inc.com}), Yiting Lu$^{1,2}$ (\url{luyiting.lyt@alibaba-inc.com}), Ru-Ling Liao$^2$ (\url{ruling.lrl@alibaba-inc.com}), Yan Ye$2$ (\url{yan.ye@alibaba-inc.com}) and Zhibo Chen$^1$ ((\url{chenzhibo@ustc.edu.cn})\\
\textbf{Affiliations:}\\
$^1$Intelligent Meida Computing Lab (IMCL). \\
$^2$Alibaba. \\

\subsection*{ECNU-SJTU VQA Team}
\textbf{Members:} Wei Sun$^1$ (\url{sunguwei@gmail.com}), Linhan Cao$^2$ (\url{caolinhan@sjtu.edu.cn}), Yuqin Cao$^2$ (\url{caoyuqin@sjtu.edu.cn}), Weixia Zhang$^2$ (\url{zwx8981@sjtu.edu.cn}), Wen Wen$^3$ (\url{wwen29-c@my.cityu.edu.hk}), Kaiwei Zhang$^2$ (\url{zhangkaiwei@sjtu.edu.cn}), Zijian Chen$^2$ (\url{zijian.chen@sjtu.edu.cn}), Fangfang Lu$^4$ (\url{lufangfang@shiep.edu.cn}), Xiongkuo Min$^2$ (\url{minxiongkuo@sjtu.edu.cn}) and Guangtao Zhai$^2$ (\url{zhaiguangtao@sjtu.edu.cn})\\
\textbf{Affiliations:}\\
$^1$East China Normal University.\\
$^2$Shanghai Jiao Tong University.\\
$^3$City University of Hong Kong. \\
$^4$Shanghai University of Electric Power.

\subsection*{HKUST-Cardiff-MI-BAAI}
\textbf{Members:} Erjia Xiao$^1$ (\url{exiao469@connect.hkust-gz.edu.cn}), Lingfeng Zhang$^2$ (\url{lfzhang715@gmail.com}), Zhenjie Su$^3$ (\url{suzhen-jie2023@cuc.edu.cn}), Hao Cheng$^1$ (\url{hcheng046@connect.hkust-gz.edu.cn}), Yu Liu$^4$ (\url{yuliu@hfut.edu.cn}), Renjing Xu$^1$ (\url{renjingxu@hkust-gz.edu.cn}), Long Chen$^5$ (\url{longchen@xiaomi.com}), Xiaoshuai Hao$6$ (\url{xshao@baai.ac.cn}) \\
\textbf{Affiliations:}\\
$^1$ The Hong Kong University of Science and Technology (Guangzhou). \\
$^2$Tsinghua University. \\
$^3$Communication University of China. \\
$^4$Hefei University of Technology. \\
$^5$Xiaomi EV. \\
$^6$Beijing Academy of Artificial Intelligence. \\

\subsection*{Media Convergence and Communication Experimental Team}
\textbf{Members:} Zhenpeng Zeng$^1$ (\url{2473910949@qq.com}), Jianqin Wu$^1$ (\url{510483263@qq.com}), Xuxu Wang$^1$ (\url{2330711901@qq.com}) and Qian Yu$^1$ (\url{1261591905@qq.com})\\
\textbf{Affiliations:}
$^1$Communication University of China. \\

\subsection*{EasyVQA}
\textbf{Members:} HuBo$^1$ (\url{hubo90@cqupt.edu.cn}) and WangWeiwei$^1$ (\url{s240231049@stu.cqupt.edu.cn}) \\
\textbf{Affiliations:}\\
$^1$Chongqing University of Posts and Telecommunications, Chongqing. \\

\subsection*{Rochester}
\textbf{Members:} Pinxin Liu$^1$ (\url{pliu23@ur.rochester.edu}), Yunlong Tang$^1$ (\url{yunlong.tang@rochester.edu}), Luchuan Song$^1$ (\url{lsong11@rochester.edu}), Jinxi He$^2$ (\url{jhe44@u.rochester.edu}) and Jiarui Wu$^2$ (\url{jiaruiwu@andrew.cmu.edu}) \\
\textbf{Affiliations:} \\
$^1$University of Rochester.\\
$^2$Carnegie Mellon University.\\

\subsection*{brucelyu17}
\textbf{Members:} Hanjia Lyu$^1$ (\url{brucelyu17@gmail.com}) \\
\textbf{Affiliations:} \\
$^1$University of Rochester.\\

\end{document}